\documentclass[letterpaper, 10 pt, conference]{ieeeconf}  

\IEEEoverridecommandlockouts                              

\usepackage[utf8]{inputenc}
\usepackage[T1]{fontenc}
\usepackage{cite}
\usepackage{amsmath,amssymb,amsfonts}
\usepackage{algorithmic}
\usepackage{graphicx}
\usepackage{textcomp}
\usepackage{xcolor}
\usepackage{breqn}
\usepackage{textcase}
\usepackage[export]{adjustbox}
\usepackage{subcaption}
\usepackage{listings}
\usepackage{hyperref}

\title{\LARGE \bf
Automatic lane change scenario extraction and generation of scenarios in OpenX format from real-world data}

\author{Dhanoop Karunakaran$^{1}$, Julie Stephany Berrio$^{1}$, Stewart Worrall$^{1}$, Eduardo Nebot$^{1}$
\thanks{
This  work  has  been  funded  by  the  Australian  Centre  for Field Robotics (ACFR), University of Sydney and Insurance Australia Group (IAG) and iMOVE CRC and supported by the  Cooperative  Research  Centres  program,  an  Australian Government initiative.
}
\thanks{$^{1}$D.Karunakaran, J.Berrio, S. Worrall,  E. Nebot  are with the Australian Centre for Field Robotics (ACFR) at the University of Sydney (NSW, Australia).
       E-mails: {\tt\small \{d.karunakaran,  j.berrio, s.worrall,  e.nebot\}@acfr.usyd.edu.au}}
}

\begin{document}

\maketitle
\thispagestyle{empty}
\pagestyle{empty}

\begin{abstract}

Autonomous Vehicles (AV)'s wide-scale deployment appears imminent despite many safety challenges yet to be resolved. The modern autonomous vehicles will undoubtedly include machine learning and probabilistic techniques that add significant complexity to the traditional verification and validation methods. Road testing is essential before the deployment, but scenarios are repeatable, and it's hard to collect challenging events. Exploring numerous, diverse and crucial scenarios is a time-consuming and expensive approach. The research community and industry have widely accepted scenario-based testing in the last few years. As it is focused directly on the relevant critical road situations, it can reduce the effort required in testing. The scenario-based testing in simulation requires the realistic behaviour of the traffic participants to assess the System Under Test (SUT). It is essential to capture the scenarios from the real world to encode the behaviour of actual traffic participants. 
This paper proposes a novel scenario extraction method to capture the lane change scenarios using point-cloud data and object tracking information. This method enables fully automatic scenario extraction compared to similar approaches in this area. The generated scenarios are represented in OpenX format to reuse them in the SUT evaluation easily. The motivation of this framework is to build a validation dataset to generate many critical concrete scenarios. The code is available online at \url{https://github.com/dkarunakaran/scenario\_extraction\_framework}

\end{abstract}

\section{Introduction}

There are six levels of automation in automobiles\cite{standard2018j3016}. For the automation levels L0, L1 and L2, the human driver has the ultimate responsibility of safety \cite{koopman2019safety,koopman2016challenges}. On the other hand, Advanced Driver Systems (ADS) are accountable for the vehicle's safety in automation levels L3, L4 and L5. In this case, it is crucial to evaluate the system in diverse challenging scenarios rather than simple test cases\cite{xinxin2020csg}. Many traditional validation approaches such as test matrix are inadequate for assessing the advanced systems efficiently due to the use of simple and repeated non-challenging scenarios\cite{zhao2016accelerated}.

Road testing is widespread and essential testing approach to validate ADS in the automobile industry\cite{waymo2017road}. However, most driving data collected during road testing is considered non-critical. Thus most of the scenarios are not relevant enough for evaluation purposes\cite{xinxin2020csg}. Besides, collected evidence will be undermined with every software update. It affects the economic, speedy, and safe aspects of AV deployment. Scenario-based testing is widely accepted in the research community and automotive industry that complement the road testing approach\cite{de2017assessment, enable2016enable,putz2017system}. This approach for validating Highly Automated Vehicles (HAVs) is promising as it reduces the test efforts by focusing on meaningful scenarios without the need of driving millions of kilometres \cite{elrofai2016scenario, amersbach2019functional, ponn2020identification}.

In scenario-based testing, generating the scenario for assessing the system is crucial. There are two types of scenario generation: knowledge-driven and data-driven\cite{hauer2020clustering,najm2007pre}. A knowledge-driven approach can generate scenarios simply and cost-effectively using expert knowledge. At the same time, a data-driven model extracts the scenarios from real-world data. Even though the data-driven approach is time-consuming, it encodes the behaviour of the real-world traffic participants that is essential for validating the ADS. 

There are different data-driven scenario extraction methods in the literature. In \cite{de2020real}, authors implemented an automated tagging in real-time and made use of a combination of tags to mine the scenarios during the second stage. For instance, when a cut-in event happens in real-time, tag such as "lane change" and "leader" is labelled for lane changing vehicle and the "following lane" for ego-vehicle. During the scenario-mining stage, they use a combination of these tags to predict a scenario. The tag combination means that a vehicle changes the lane and becomes a leader, and the ego vehicle follows the vehicle. The method is highly dependent on real-time tagging and is not easy to implement to extract scenarios using other published datasets. \cite{krajewski2018highd} published an extensive scale vehicle trajectory data based on German highways called highD and demonstrated the lane-change scenario extraction by detecting lane changes using lane crossing. The extraction method is useful in image-based birds-eye view datasets and not beneficial to other types of datasets. \cite{xinxin2020csg} enables the critical concrete scenario generation in simulation from real-world accident video data. One of the significant components of the method is the scenario extraction from video data using various computer vision techniques. In \cite{tenbrock2021conscend}, authors introduce a methodology to generate concrete scenarios by extracting scenario parameters from a highD dataset for assessing the Active Lane Keeping System (ALKS). OpenSCENARIO and OpenDRIVE format describe the generated concrete scenarios and road network. \cite{xinxin2020csg} and \cite{tenbrock2021conscend} has proposed scenario extraction frameworks and generated scenarios represent in common format. Both approaches use mainly image-based methods, and adverse weather conditions can impact the performance of the extraction method. Also,  both works required manual creation of OpenDRIVE file using MATLAB's roadrunner, which makes the methods not fully automated.

National Highway Traffic Safety Administration (NHTSA) conducted an extensive study on the crash scenarios based on the 2004 General Estimates System (GES) crash database\cite{najm2007pre}. They have found that the lane change scenario accounts for 7.62$\%$ of Two-Vehicle Light-Vehicle crashes, equivalent to 295,000 crashes. As scenario-based testing aims for meaningful scenarios to assess the SUT, extracting the lane-change scenario from real-world data is essential. This paper focuses on a scenario extraction framework that can extract the variety of lane change scenarios such as cut-in and cut-out.

This paper proposes a novel data-driven scenario extraction framework for capturing the lane-change scenarios from real-world data. It uses lidar data to extract the road networks. Then uses provided tracking information and road network to identify and extract the lane change scenarios. The extracted scenario and road networks are represented in OpenX formats such as OpenSCENARIO and OpenDRIVE. The proposed approach is a lidar-based method that provides more robust results in diverse environmental conditions than many image-based techniques. The framework is an automated scenario extraction pipeline that generates OpenX files from real-world data without any manual intervention required compared to similar works \cite{xinxin2020csg,tenbrock2021conscend}. The extraction of the scenario means we are extracting parameters from the real-world scenario. Then we use these parameters to generate the scenarios in simulation. 

The framework enables us to extract the parameters from real-world data. It helps build parameter space/logical scenarios to generate many concrete scenarios using stochastic variation methods. Then we can enable these concrete scenarios to validate SUT. Typical logical scenarios will have a range or distribution for each parameter. In the extended work, we build a validation dataset that includes logical scenarios. The paper's primary emphasis is the scenario extraction pipeline, and we demonstrate the framework's effectiveness by comparing simulated and real-world trajectories. 


\section{Background}

Real-world data for autonomous vehicles testing is collected using various sensors like cameras, lidar, IMU, etc., mounted in a mobile platform. This data is processed and saved in a format accessible by the different perception, planning, control and localisation algorithms. This section presents the selection of the data format for our algorithm, the data collection platform, the coordinate frames used, and the type of extracted scenarios.

\subsection{Open format}

The format of the testing scenarios should be readable by a software to reconstruct or simulate the real environment.
Open Standard is a collective term that is used to refer to OpenDRIVE \cite{opendrive} and OpenSCENARIO \cite{openscenario}, three approaches that were started at the grassroots level in the industry to standardize how road data is described and formatted \cite{openstandar}.
The OpenDRIVE is an XML format that describes the road network\cite{opendrive}. Creating the OpenDRIVE file can recreate the road network and content in any supported simulation to verify and validate autonomous vehicles.

Furthermore, OpenSCENARIO describes the dynamic contents such as the behaviour of the traffic participants and weather conditions\cite{openscenario}. OpenScenario describes the complex maneuvers of the multiple road participants for verification and validation of autonomous driving in simulation. It is also represented as an XML, and compatible with different simulator software. In this paper, the road and the description of the events are depicted using these two open formats.

\begin{figure}[h]
    \centering
    \includegraphics[width=0.95\columnwidth]{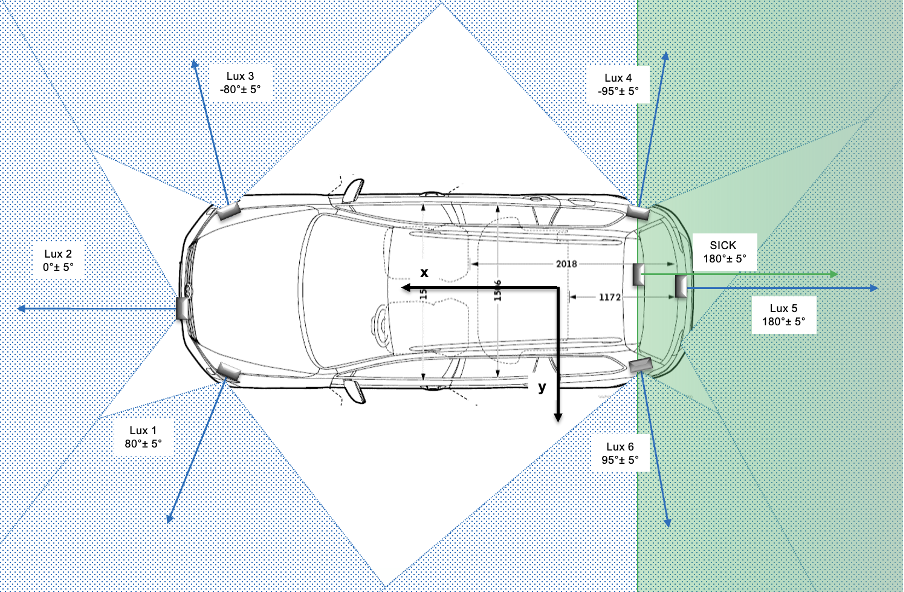}
    \caption{\small Top-down view of the distribution of the data collection vehicle's lidar sensors. In blue is the field of view for the ibeo LUX lidar and in green is the FOV of the SICK lidar. }
    \label{fig:data_collection_vehicle}
\end{figure}


\begin{figure}[h]
\centering
\includegraphics[width=0.95\columnwidth]{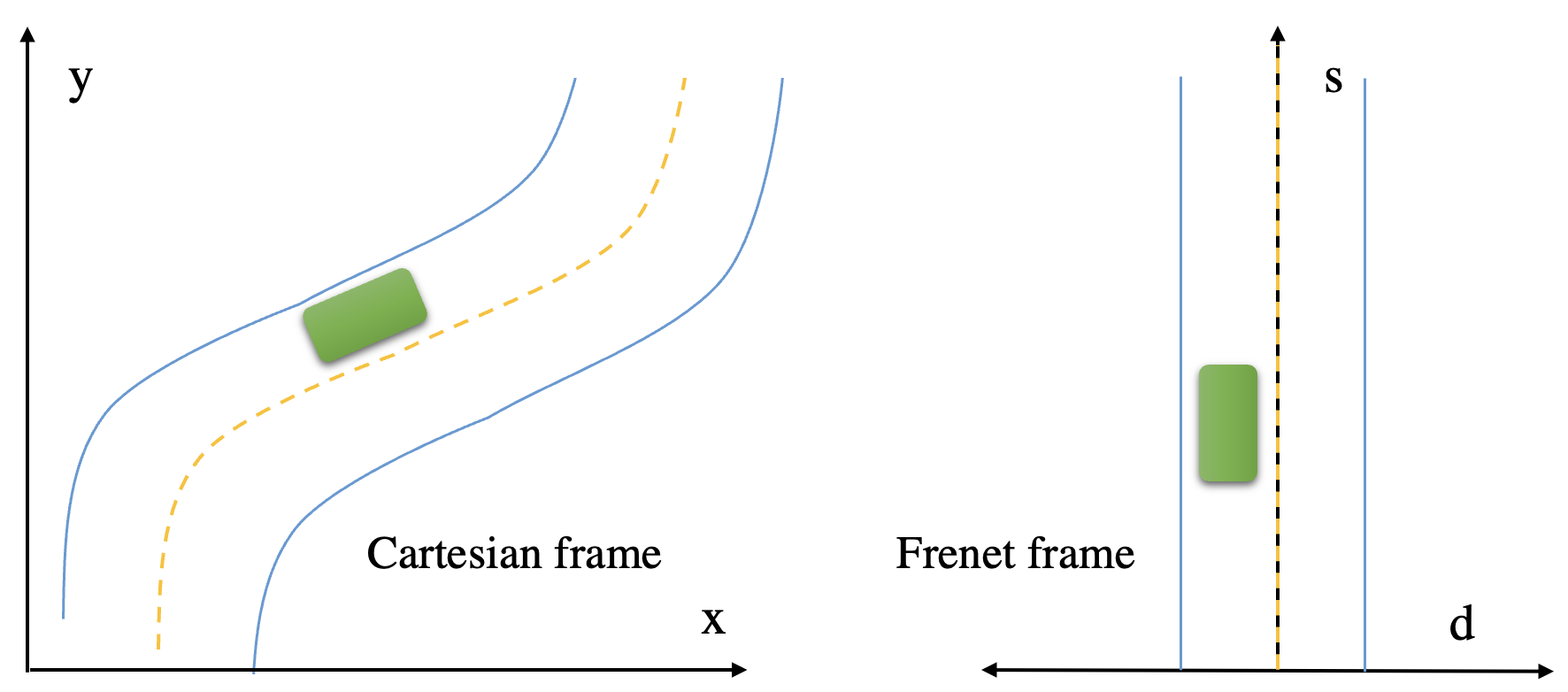}
\caption{\small Catersian frame vs Frenet frame }
\label{fig:cart_to_frenet}
\end{figure}

\subsection{Data collection vehicle}

The data collection vehicle is a Volkswagen Passat station wagon fitted with Ibeo and SICK lidars. Ibeo Automotive Systems GmbH provides HAD Feature Fusion detection and tracking system to track the road participants. The system comprises 6 Ibeo LUX 4 beam (25 Hz lidar scanners) and an on-board computer to detect and track the objects up to 200 meters in real-time. One SICK lidar is fitted to the top of the roof that points to the car's rear side. The placement of the lidars on the vehicle is shown in the figure~\ref{fig:data_collection_vehicle}. 

In our work, we use the data from these sensors to extract the road network. The tracking data contains real-time estimates of the road users such as user id, type, relative positioning and velocity. Also, odometry data is calculated using the IMU retrofitted in the vehicle.

Our framework requires tracking, IMU and lidar point cloud data. It is not dependent on a specific sensor suite and can extract the scenarios with minimal modification to the code when the input requirements are met.

\subsection{Coordinate frames}

In this paper, we use \textit{base\_link} and \textit{odom} frames to reference the vehicle's position. The origin of the \textit{base\_link} frame is at the centre of the robot. The \textit{base\_link} frame changes when the robot position changes. The \textit{odom} frame's origin is at the starting location of the platform. Moreover, this frame is fixed in the world.

In a cartesian coordinate frame ($x$, $y$), location represents the position of the road participants. It is not an adequate representation of the curvy roads. Frenet Coordinates represent the vehicle's location on the road more intuitively than cartesian coordinates; thus, it is mathematically easier to describe the typical behaviour of the traffic participants. There are two coordinates for the frame: $s$(longitudinal displacement) and $t$(lateral displacement). The $s$ coordinates define the distance along the road, or reference line and $t$ represents the perpendicular distance from each $s$ position to the centre of the road or reference line\cite{openscenario, werling2010optimal}.

\subsection{Lane change scenario types}

\begin{figure}[t]
\vspace{2mm}
    \centering
    \begin{subfigure}[b]{0.47\columnwidth}
         \centering
         \includegraphics[width=0.8\columnwidth]{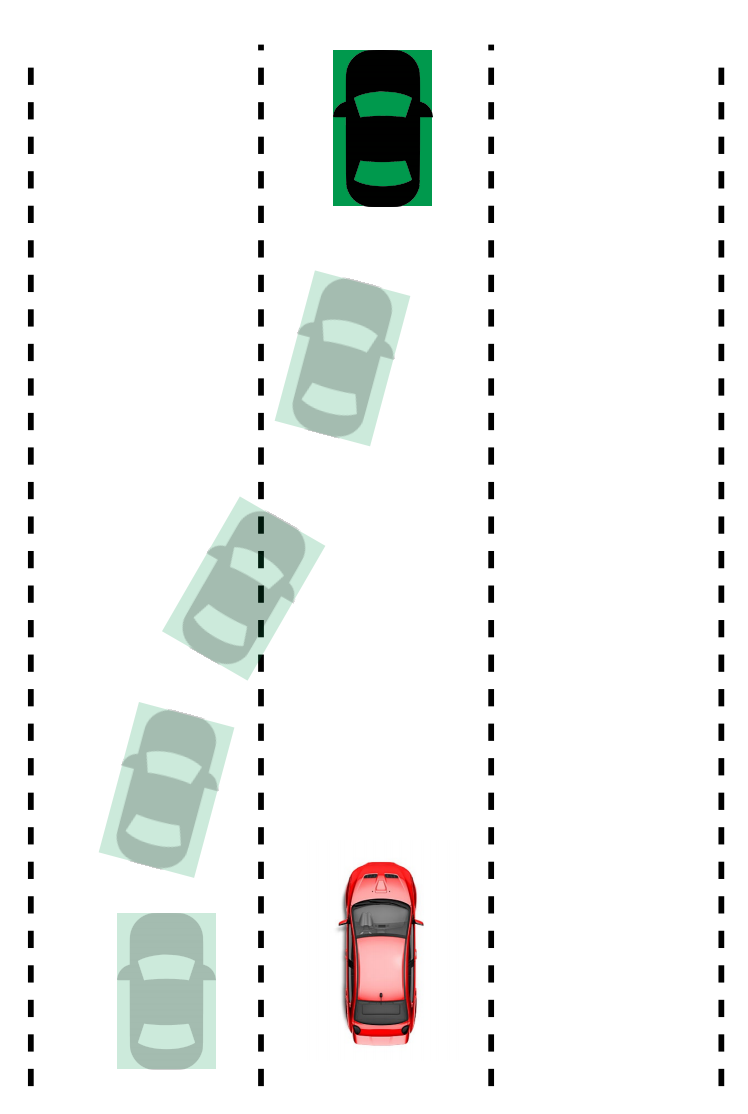}
         \caption{Cut-in scenario: the vehicle(green) that changes the lane to ego vehicle's lane(red) }
         \label{fig:cut_in}
    \end{subfigure}
    \hfill
    \begin{subfigure}[b]{0.47\columnwidth}
         \centering
         \includegraphics[width=0.8\columnwidth]{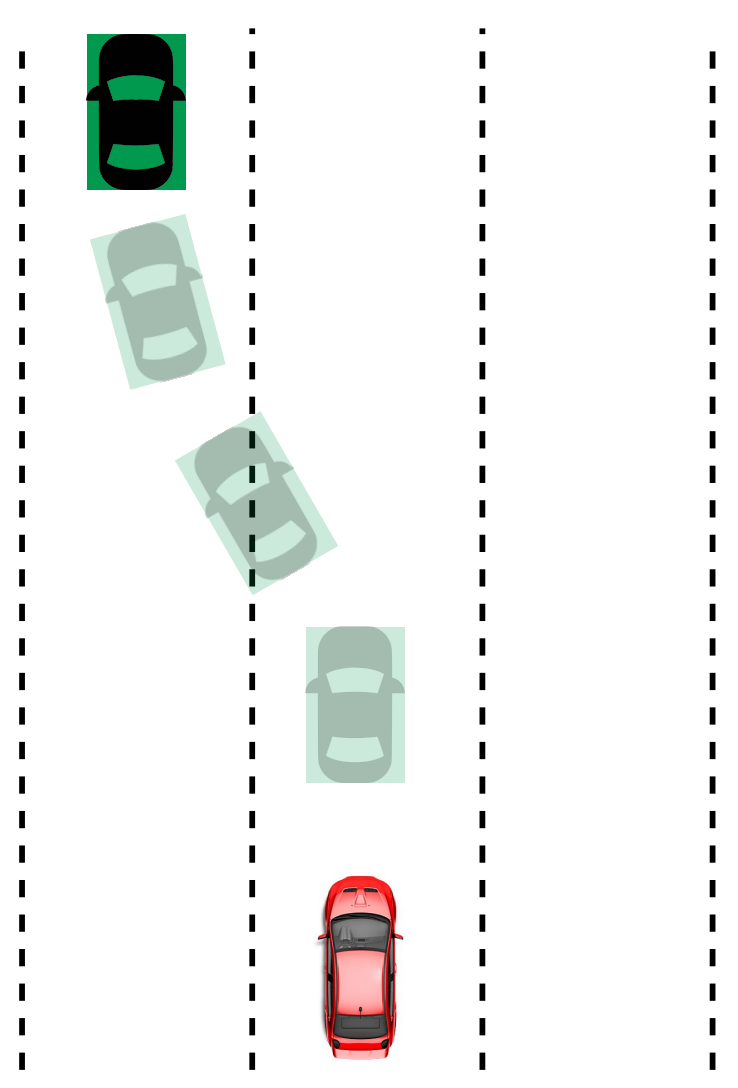}
         \caption{Cut-out scenario: the vehicle(green) that changes the lane from ego vehicle's lane(red) }
         \label{fig:cut_out}
    \end{subfigure}
    \caption{Lane change scenario types}
    \label{fig:lane_change_type}
\end{figure}

As shown in the figure~\ref{fig:lane_change_type}, there are two lane change scenarios: cut-in and cut-out. When the vehicle moves into the ego-vehicle lane, it is considered as cut-in scenario (figure~\ref{fig:cut_in}). Contrary,  cut-out scenario (figure~\ref{fig:cut_out}) happens when the front vehicle changes the lane.

\section{Lane change scenario extraction framework}

\begin{figure}[t]
\vspace{2mm}
\centering
\includegraphics[width=0.9\columnwidth]{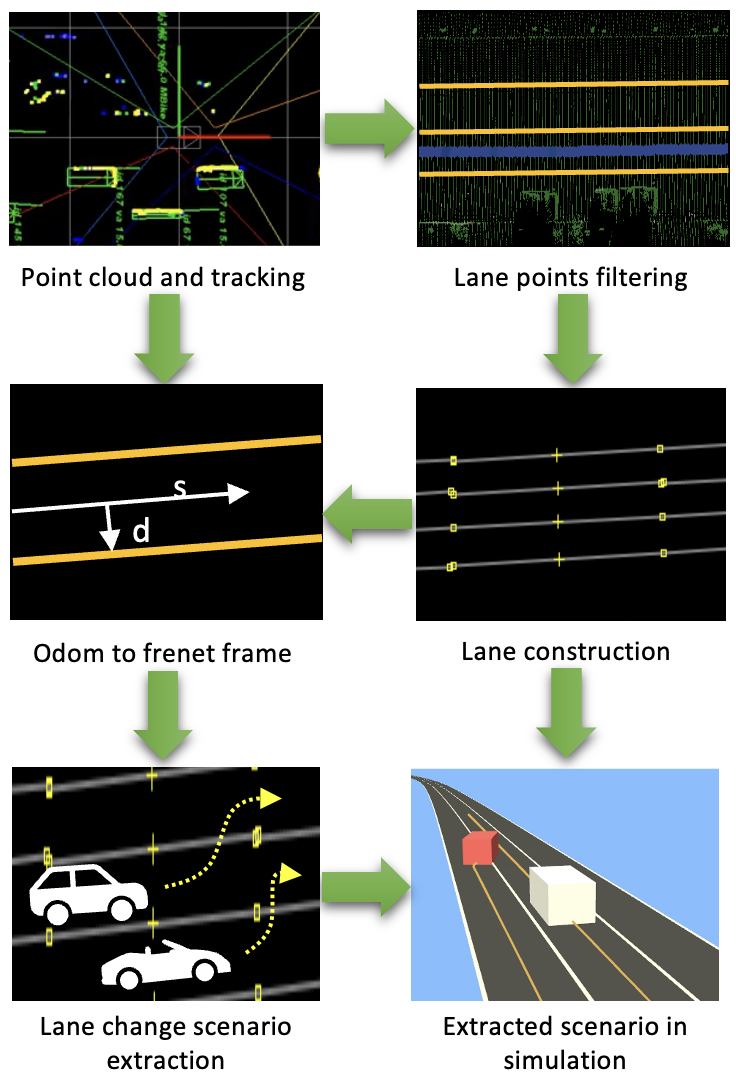}
\caption{\small Pipeline of scenario extraction framework: from data obtained by the collection system, lane change scenarios are identified and parameterized in open format. }
\label{fig:framework}
\end{figure}

As depicted in figure~\ref{fig:framework}, the proposed framework takes the point cloud, IMU, and objects tracking data as input. It outputs the extracted scenarios in OpenSCENSRIO format and the corresponding road network in the OpenDRIVE format. The framework has three parts: parameter extraction, OpenX files generation and simulation. 

\subsection{Data}

The real-world data is stored in rosbag format to play it back in the ROS environment. The framework uses odometry, point cloud, and object tracking data to detect and extract the lane change scenarios. Moreover, the framework can extract the scenarios using any system that provides IMU, the point cloud with intensity values and vehicle tracking data with minimal modification. 

We use the odometry data to estimate the vehicle’s relative position from a starting location, hence, the position of the ego vehicle. 
The data collection vehicle is equipped with an object tracking system that captures the road participants trajectory in real-time. We use the \textit{TF} transformation in ROS to convert the position in \textit{base\_link} frame to the \textit{odom} frame, which gives a fixed position in the world.

\subsection{Parameter Extraction}

The first part of the framework identifies the lane change scenarios from the real-world data and extracts the scenarios parameters. These parameters are used to construct the OpenSCENARIO files that emulate real-world lane change situations.

The main components of the pipeline are explained below.

It is essential to know the ego vehicle's location and other road participants within a common frame to detect cut-in and cut-out events. In this case, we rely on the odometry and lidar data to create a lane map that describes the road. We have applied various techniques to extract the lidar points on the lane markings. Initially, all the lidar points are sorted from the centre to its sides. Then, to extract only the points hitting the road, we evaluate the first and second derivative of the angle between adjacent points. In this way, we can identify the curbs and separate road and non-road points. To identify points on lane marking, we use the intensity information provided by the laser in the point cloud.
The lane markings are usually drawn with reflective paint, so their intensity is relatively higher than other road points. We extract the points with high intensity and consider them as lane points.

\begin{figure}[t]
\vspace{2mm}
\centering
\includegraphics[width=0.85\columnwidth]{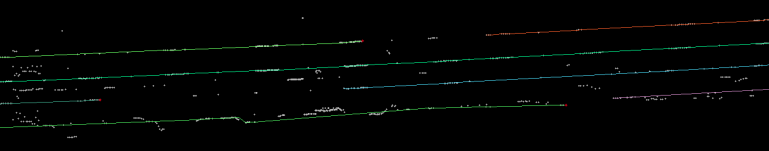}
\caption{\small Lane construction from pointcloud.}
\label{fig:connected_lanes}
\end{figure}

 
\textbf{Lane construction}: We maintain two separate stacks for constructing the lanes, one for lane-mark points clusters and a second for clusters belonging to the same lane. The first task is to cluster the points that belong to the same lane marking. After filtering out the lane points in each lidar scan, these lane points are transformed to the Odom frame, and then we loop through them to identify which of the existing clusters in the first stack is closer.  If the distance between the new lane point and the nearest point from the closest cluster is less than one meter, the new lane point is added to this cluster. Otherwise, the new lane point will be added as a new cluster in the stack and set the earlier clusters as inactive. We increase the inactive count by 1 for the clusters far from 
new lane points. The inactive count is reset to zero when a new lane point is added. Once we have all the lane points of a lane marking in a cluster, we move this group of lane points from the first stack to the second stack. We use the inactivity counter to decide when to move the clusters to the second stack, as shown in  equation~\ref{eq2}. 
\begin{equation}
\label{eq2}
\Bigg\{\begin{matrix}
move & inactive\ count > 20\ lidar\ scans   \\
Not\ move & otherwise
\end{matrix}
\end{equation}
The threshold for inactive\_count is selected by the trial and error method.

We take the first and the last point for each new and complete cluster in the first stack to create a line segment. Then, we loop through all the clusters in the second stack to identify which cluster can be merged with the new line segment. To do so, we find the point corresponding to the projection of the existing clusters line into the new line and compute the distance between the projected point and the first point of the new line segment.

The projected point of the existing clusters is calculated using their slope-intercept form of a line. We take the last line segment added to the cluster and compute its slope and intercept constants. Then we calculate $y$ for the corresponding $x$ value taken from the new line's first point. The $x$ and $y$ coordinates give the projected point of a cluster into the new line. If the Euclidean distance between the projected point of the nearest cluster and the starting point of the new line segment is less than 0.25 m, then the new line segment is added to the closest cluster. The equation~\ref{eq4} defines the merging criteria. If the line segment is not added to the cluster, we generate a new lane cluster with the line segment and add it to the second stack.
\begin{equation}
\label{eq4}
\Bigg\{\begin{matrix}
merge & distance < 0.25   \\
not\ merge & otherwise
\end{matrix}
\end{equation}

In a straight road, the computed distance is always closer to zero, but this distance is greater in the case of curvy roads and missing lane markings. We found that 0.25 m value as a distance threshold fit through the experimentation in most cases. However, this threshold may not work well with sharp curvy roads, but it can be tuned depending on the case.
This whole process repeats until we finish reading the lidar data. The figure~\ref{fig:connected_lanes} depicts the constructed lane. We construct the lines from the line segments, and the collection of such lines forms the lanes.

\begin{figure}[t]
\vspace{2mm}
\centering
\includegraphics[width=0.85\columnwidth]{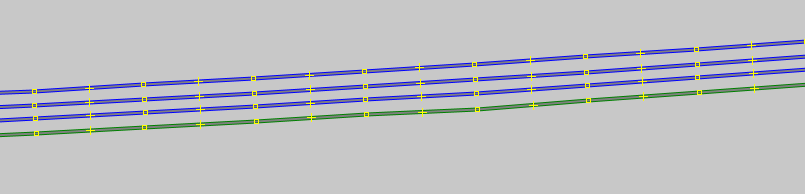}
\caption{\small The generated Lanelet map from constructed lanes from the pointcloud.}
\label{fig:laneletmap}
\end{figure}

In some cases, the lane construction stage creates broken lanes due to intersection and missing lane markings. For instance, if a portion of the middle lane is missing, we make use of an intermediate representation called lanelets\cite{poggenhans2018lanelet2} to complete them. In the lanelets, there are three essential elements: points, linestrings, and lanelets. The linestring is a collection of points, and the lanelet is formed using the left and right linestrings. An example of lanelet map representation is shown in the figure~\ref{fig:laneletmap}. We divide the entire road into twenty-five-meter sections and create a LineString for each lane in the section. If the distance between two LineStrings is more than 5 meters, then that indicates the missing lane. We create a linestring between existing LineStrings to fill the missing lane. We can tune the length of the section depending on the use case. In many cases, the length between 20 to 25 meters works well for filling the missing lanes.

\begin{figure}[t]
\vspace{2mm}
\centering
\includegraphics[width=0.85\columnwidth]{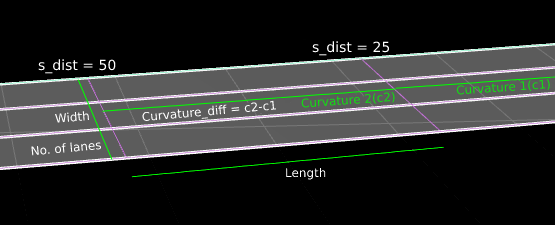}
\caption{\small Parameters for creating the OpenDRIVE file. The road is divided into many sections, and at each section, the parameters are computed.}
\label{fig:line_to_od}
\end{figure}

As discussed earlier, we divide the whole road into twenty-five-meter segments/sections. At each section, we compute five parameters for generating OpenDRIVE file shown in the figure~\ref{fig:line_to_od}: length, width, no\_of\_lanes, curvature\_diff, and s\_dist. We compute the average width of the lane and keep the section's length. In this experiment, the length of the section is 25 meters. The road curvature is determined by the relative heading between two road segments. We use the curvature\_diff parameter to find the relative heading. The parameter is the difference between the previous segment's curvature and the current curvature. If the curvature difference is zero, both sections have relatively no heading changes. The curvature of each section is computed by taking the gradient of the reference line(centre line) of each section. The s\_dist is the distance travelled along the road or reference line. This parameter value is similar to longitudinal displacement coordinates in the Frenet frame.

\textbf{Odom to Frenet frame}: the Frenet frame has two coordinates: longitudinal displacement, \textit{s} and lateral displacement, \textit{t}. We use the ego path as the reference line to compute \textit{s} coordinate. At this stage, the vehicle's position is in the \textit{odom} frame. The \textit{t} is the perpendicular distance from the vehicles' position to the reference line. We use vehicles' positions in the Frenet frame and lane number to detect the lane change scenarios. For instance, the lateral displacement of the vehicle that is likely to exercise the cut-in scenario will always be greater than 1.5 meters to the reference line (ego path). 


\textbf{Parameter extraction logic}: Lane change scenarios are detected by comparing vehicles' lane position  and their lateral distance to the ego path at each second.

For cut-in events, we keep track of all vehicles on a different lane than the ego vehicle's lane and vehicles' lateral displacement from the ego path greater than 1.5 meters away. If we consider the lane's width as 3 meters, then 1.5 meters away is considered another lane. We choose the lateral displacement of 1.5 meters and lane number to track the vehicle to ensure they are on different lanes. We consider only those vehicles in front of the ego vehicle or riding parallel to it. When any of these tracked vehicles' lanes become the same as the ego lane and lateral displacement is less than 0.5 meters, we mark it as cut-in scenario detection. Likewise, a cut-out scenario is detected when a vehicle moves away from the ego's lane; we keep track of all vehicles on the ego vehicle's lane, and later displacement is less than 0.5 meters. The cut-out scenario is detected and marked when any tracked vehicle changes its lane and is greater than 1.5 meters laterally. The lateral displacement, 0.5 meters, ensures that the vehicle is in the same lane as the ego vehicle.

Once we detected and marked the cut-in and cut-out scenarios, we fixed the starting timestep of the scenario as eight-second prior to the cut-in/cut-out action and the end timestep of the scenario as 5 seconds after the action.

For all the extracted scenarios we compute all the parameters mentioned in table~\ref{table:table_cut_in_cut_out} in the final step. We extract the parameters from data between start and end times noted in the marked scenario.
We have identified list of parameters in the table~\ref{table:table_cut_in_cut_out} needed to generate cut-in and cut-out scenarios in OpenSCENARIO format from real-world data.

\begin{table}[h!]
\centering
 \caption{List of parameters for cut-in and cut-out scenario}
 \label{table:table_cut_in_cut_out}
 \begin{tabular}{|c | c | c|} 
 \hline
 Parameter & Vehicle & Unit\\
 \hline
 initial_speed  & ego-vehicle & m/s\\ 
 \hline
 initial_position & ego-vehicle & m\\ 
 \hline
 initial_lane  & ego-vehicle & -1 to -4\\ 
 \hline
 speed 1 & ego-vehicle & m/s\\
 \hline
 ... & ... & ...\\
 \hline
 speed m & ego-vehicle & m/s\\
 \hline
 distance 1 & ego-vehicle & m\\
 \hline
 ... & ... & ...\\
 \hline
 distance m & ego-vehicle & m/s\\
 \hline
 triggering_distance & both & m\\
 \hline
 initial_speed & adversary & m/s\\ 
 \hline
 initial_position & adversary & m\\ 
 \hline
 initial_lane & adversary & -1 to -4\\ 
 \hline
 speed 1 & adversary & m/s\\
 \hline
 ... & ... & ...\\
 \hline
 speed m & adversary & m/s\\
 \hline
 distance 1 & adversary & m\\
 \hline
 ... & ... & ...\\
 \hline
 distance m & ego-vehicle & m/s\\
 final_lane  & adversary & -1 to -4\\ 
 \hline
 \end{tabular}
\end{table}

We extract initial speed, position, and lane parameters for ego and adversary vehicles. There is a $m$ number of speed and distance parameters for both vehicles. We capture speed and distance travelled at $m$ number of samples in an overall scenario. For instance, the first sample is the start of the scenario, where the end of the scenario is the last sample.  The triggering\_distance parameter gives the required relative distance between ego-vehicle and adversary vehicle to initiate the lane change.
Currently, the extraction pipeline is not focused on complex intersections, sharpe curvy roads, and merging lanes.

\subsection{OpenX file generation}
The second stage of the scenario extraction framework focuses on creating the OpenX files from the extracted scenario parameters. The trajectories are represented in OpenSCENARIO format and road network in OpenDRIVE format.

\textbf{OpenSCENARIO file generation}: The vehicles' behaviour is defined in the OpenSCENARIO format. We first initialize the scenario objects. Our experiment has two road participants: ego and adversary vehicle. We use the extracted parameters such as initial velocity, position, and lane parameters to define the objects' initial behaviour.

We use condition and action to create an event in OpenSCENARIO. For instance, a cut-in/cut-out event for the adversary vehicle is created using relative distance conditions and lane change action. We use the triggering\_distance parameter for the condition and the final\_lane parameter for the lane change action. In order to do the lane change action, the vehicles have to follow similar trajectories as in real-world data to meet the above condition. We have used speed and distance parameters to create a similar trajectory. In order to do that, we create $m$ events with travelled distance conditions and speed action. For each of these events, the distance parameter is used as a condition, and once the condition is met, we initiate the action using the corresponding speed parameter.

\begin{figure}[t]
\vspace{2mm}
\centering
\includegraphics[width=0.85\columnwidth]{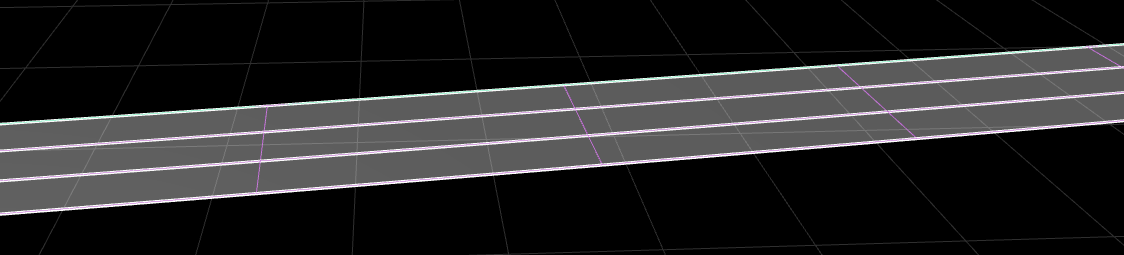}
\caption{\small Generated OpenDRIVE file from the details stored per section.}
\label{fig:opendrive}
\end{figure}

\textbf{OpenDRIVE file generation}: OpenSCENARIO only defines the dynamic content. In order to run the dynamic content, OpenSCENARIO requires reference to OpenDRIVE that defines the road networks. In this paper, we automate this process to make the scenario generation pipeline fully automated without manually creating the OpenDRIVE file as shown in the figure~\ref{fig:opendrive}. The whole process does not require manual intervention to create or modify any process compared to similar work in this area\cite{xinxin2020csg,tenbrock2021conscend}.

We have captured five parameters, as shown in the figure~\ref{fig:line_to_od} in each section of the road. We start with defining the planView element in OpenDrive. It defines the shape of the road. As we create the road network section by section, we need to define the shape of each section. We use curvature\_diff of the current and previous sections to create a spiral shape. We create this shape for each section and add it to the planView element. The next task is to define lanes of the road using lanes element. We use the laneSection element to create the number of lanes per section. As plainView defines the shape of the section, we create the same number of laneSection elements as planView elements. The laneSection requires s\_dist parameter and the number of lanes in each road section from the no\_of\_lanes parameter. In the end, we add planView and laneSection elements to the road element that creates the whole structure of the OpenDRIVE file. 

\subsection{Simulation}

The extracted scenario is represented in OpenX format. The generated scenario can be run in any simulation that supports the OpenX format. We use OpenSCENARIO player, Esmini to demonstrate and do a similarity check. The Carla simulator has a scenario running tool that supports the OpenX format. The file generated is well suited for Carla\cite{dosovitskiy2017carla} as well. 

\section{Results}

The results show that the proposed architecture enables lane change scenario extraction from real-world data to OpenX file generation. Using our framework, we can build a validation dataset with numerous lane change scenarios extracted from real-world data to evaluate SUT.

\begin{figure}
\vspace{2mm}
    \centering
    \hfill
    \begin{subfigure}[b]{0.32\columnwidth}
         \centering
         \includegraphics[width=0.95\columnwidth]{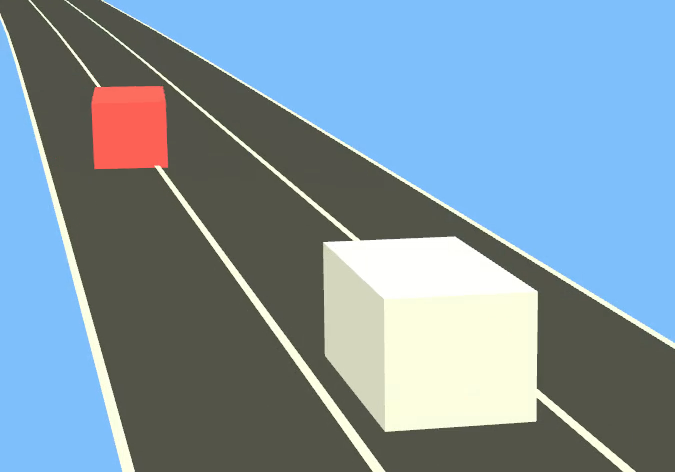}
         \caption{}
         \label{fig:cutin_example4}
    \end{subfigure}
    \hfill
    \begin{subfigure}[b]{0.32\columnwidth}
         \centering
         \includegraphics[width=0.95\columnwidth]{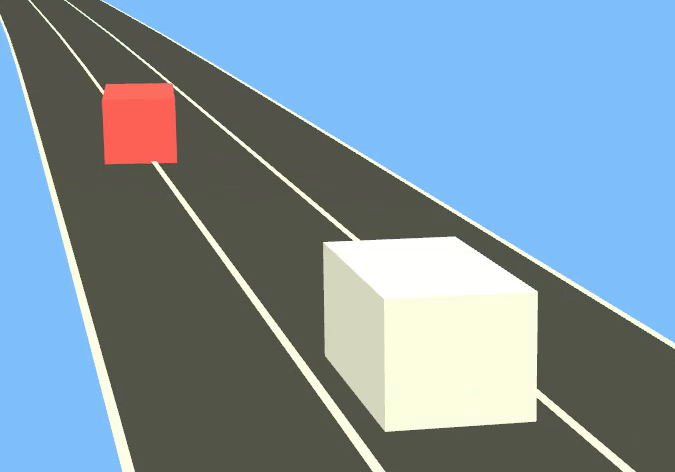}
         \caption{}
         \label{fig:cutin_example5}
    \end{subfigure}
    \hfill
    \begin{subfigure}[b]{0.32\columnwidth}
         \centering
         \includegraphics[width=0.95\columnwidth]{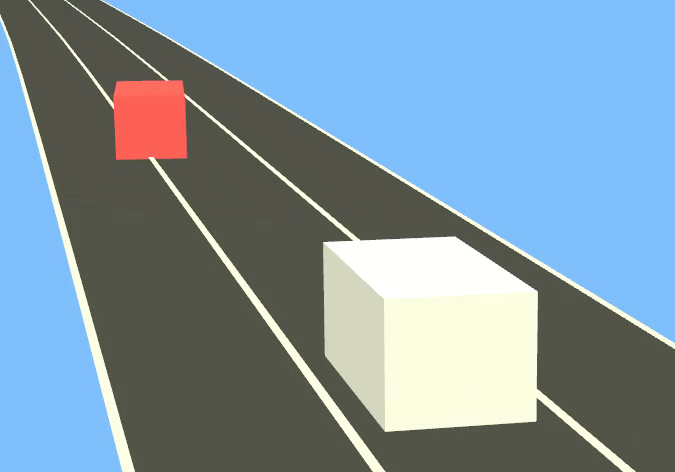}
         \caption{}
         \label{fig:cutin_example6}
    \end{subfigure}
    \hfill
    \begin{subfigure}[b]{0.32\columnwidth}
         \centering
         \includegraphics[width=0.95\columnwidth]{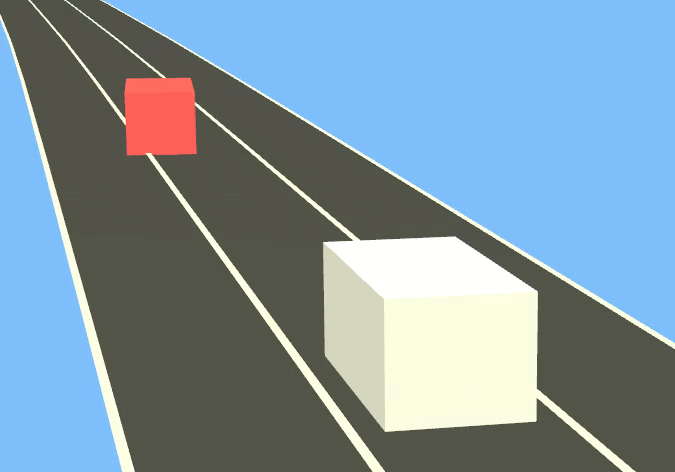}
         \caption{}
         \label{fig:cutin_example4}
    \end{subfigure}
    \hfill
    \begin{subfigure}[b]{0.32\columnwidth}
         \centering
         \includegraphics[width=0.95\columnwidth]{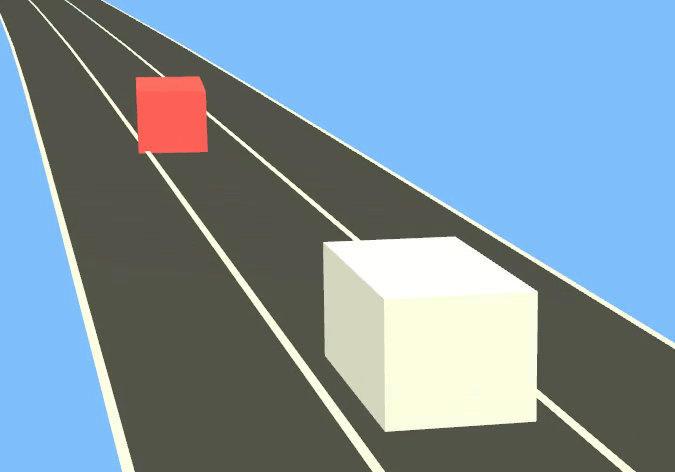}
         \caption{}
         \label{fig:cutin_example5}
    \end{subfigure}
    \hfill
    \begin{subfigure}[b]{0.32\columnwidth}
         \centering
         \includegraphics[width=0.95\columnwidth]{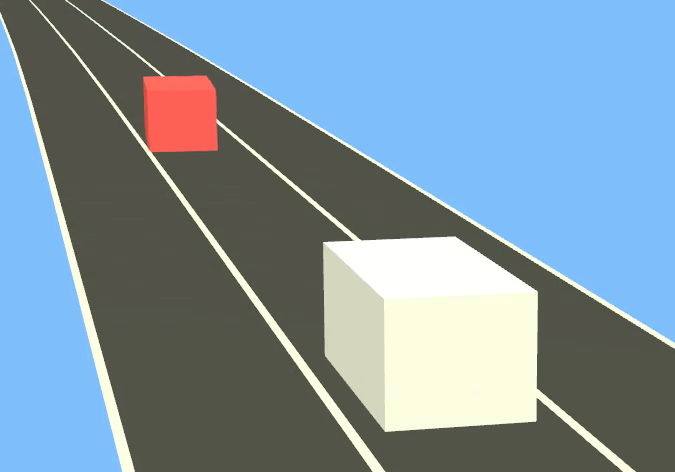}
         \caption{}
         \label{fig:cutin_example6}
    \end{subfigure}
    \hfill
    \begin{subfigure}[b]{0.32\columnwidth}
         \centering
         \includegraphics[width=0.95\columnwidth]{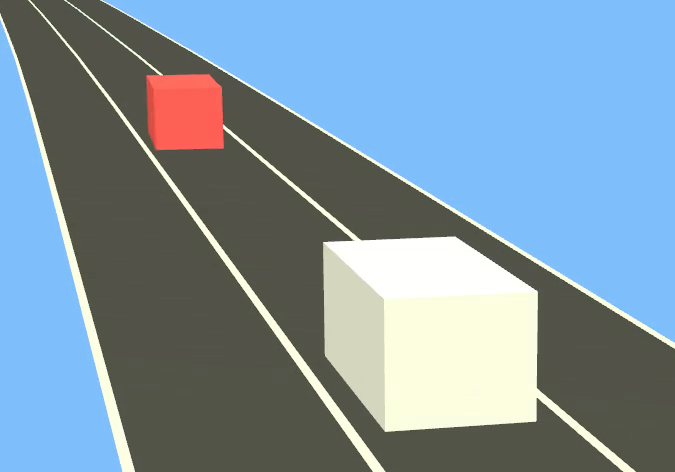}
         \caption{}
         \label{fig:cutin_example4}
    \end{subfigure}
    \hfill
    \begin{subfigure}[b]{0.32\columnwidth}
         \centering
         \includegraphics[width=0.95\columnwidth]{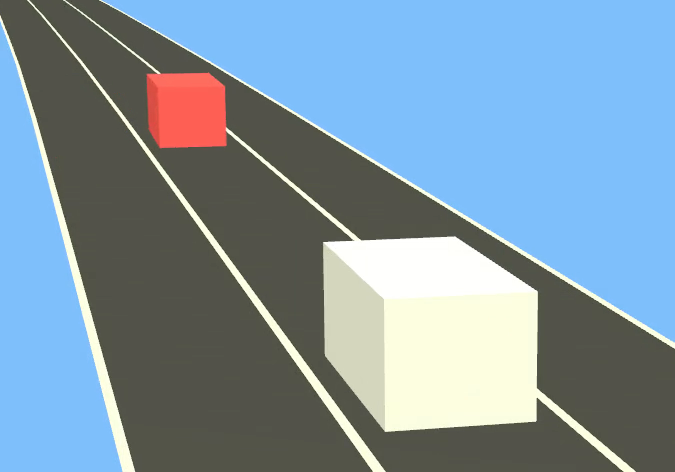}
         \caption{}
         \label{fig:cutin_example5}
    \end{subfigure}
    \hfill
    \begin{subfigure}[b]{0.32\columnwidth}
         \centering
         \includegraphics[width=0.95\columnwidth]{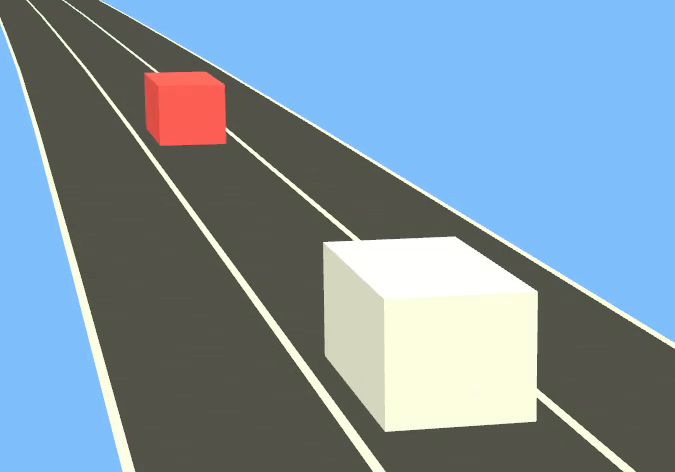}
         \caption{}
         \label{fig:cutin_example6}
    \end{subfigure}
    \caption{One example of generated cut-in scenario}
    \label{fig:cutin_example_flow}
\end{figure}

The data is collected by the vehicle and stored into rosbags. The pipeline takes as input point cloud and objects tracking information to extract the scenario parameters. Then we create OpenX files from the extracted parameters and lane information. The generated OpenSCENARIO file can play in an OpenSCENARIO player like Esmini or Carla's scenario runner. The figure~\ref{fig:cutin_example_flow} shows the generated cut-in scenarios played on OpenSCENARIO player, Esmini player.

\begin{figure}[t]

\centering
\includegraphics[width=0.95\columnwidth]{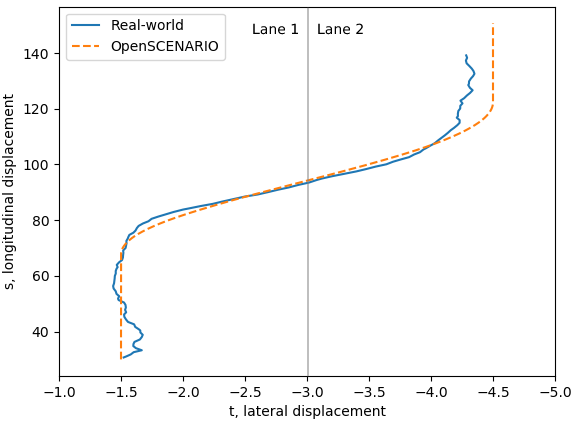}
\caption{\small It is a similarity check of the adversary vehicle's trajectory between real-world example and generated one. The figure shows that the framework can generate a similar simulated trajectory of the real-world version. The lateral displacement shown here is in negative values, and it has no impact on the result.}
\label{fig:similarity_check_pos}
\end{figure}

\begin{figure}[t]
\vspace{2mm}
\centering
\includegraphics[width=0.95\columnwidth]{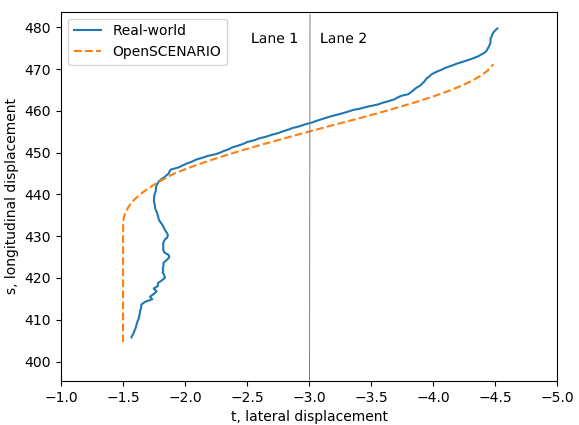}
\caption{\small It is an adversary's trajectory of cut-in scenario example. The figure depicts the similarity check of trajectories from real-world data and its simulated version. The lateral displacement shown here is in negative values, and it has no impact on the result.}
\label{fig:similarity_check_pos1}
\end{figure}

It is essential to generate scenarios similar to real-world data to encode the realistic behaviour of the traffic participants. We use the trajectory information to compare generated scenarios with real-world scenarios. We capture the lateral and longitudinal displacement from the real-world data to create the trajectory at each timestep. The thick line in the figure~\ref{fig:similarity_check_pos} shows the real-world trajectory of the adversary vehicle observed by the ego vehicle. We use the Esmini simulator to run the OpenSCENARIO file generated by the framework. In this comparison, we capture the displacement data longitudinally and laterally at each time step in the simulator to create the adversary vehicle's trajectory as shown in the dashed line in the figure~\ref{fig:similarity_check_pos}. Then we compare both real-world and simulated trajectories in the figure~\ref{fig:similarity_check_pos}. The scenario depicted in the figure has similar trajectories, indicating that the framework can regenerate real-world lane change scenarios in a simulated environment. We have shown another cut-in scenario result in the figure~\ref{fig:similarity_check_pos1}. The figure depicts the similarity check between real-world and generated trajectories. In this experiment, we do not consider lateral displacement of the vehicles as parameters that lead to a minor deviation in the vehicle's lateral position. In the extended work, we look at different parameters, including lateral displacement, to further accurately recreate the trajectory and build the validation dataset.


\begin{figure}[t]
\vspace{2mm}
    \centering
    \hfill
    \begin{subfigure}[b]{0.49\columnwidth}
         \centering
         \includegraphics[width=0.95\columnwidth]{images/trajectory_all.png}
         \caption{The figure is a trajectory of an adversary vehicle in a cut-in scenario. The total duration of the scenario is ten seconds, and we have used ten samples of speed and distance values.  The figure depicts that framework can generate simulated trajectory as close to a real-world example.}
         \label{fig:pos_10_points}
    \end{subfigure}
    \hfill
    \begin{subfigure}[b]{0.49\columnwidth}
         \centering
         \includegraphics[width=0.95\columnwidth]{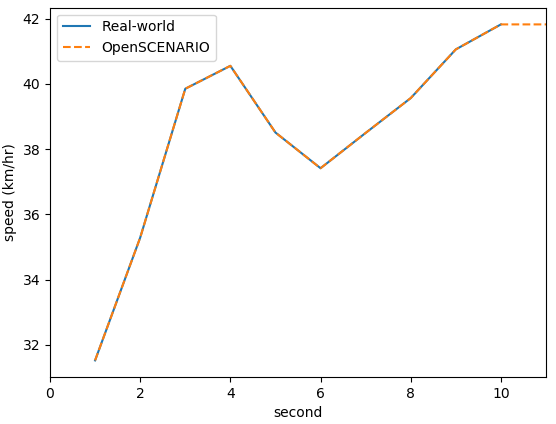}
         \caption{The figure depicts the adversary's speed profile of the same cut-in scenario shown in figure~\ref{fig:pos_10_points}. It indicates that velocity sampling is done at each second to create the trajectory. The figure shows that the simulated velocity profile is the same as a real-world example.}
         \label{fig:speed_10_points}
    \end{subfigure}
    \hfill
    \begin{subfigure}[b]{0.49\columnwidth}
         \centering
         \includegraphics[width=0.95\columnwidth]{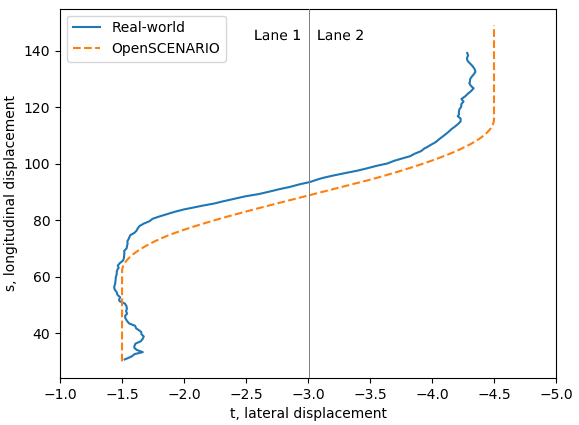}
         \caption{In this figure, we have used five samples of speed and distance values of the same cut-in scenario to create the trajectory for the adversary vehicle.}
         \label{fig:pos_5_points}
    \end{subfigure}
    \hfill
    \begin{subfigure}[b]{0.49\columnwidth}
         \centering
         \includegraphics[width=0.95\columnwidth]{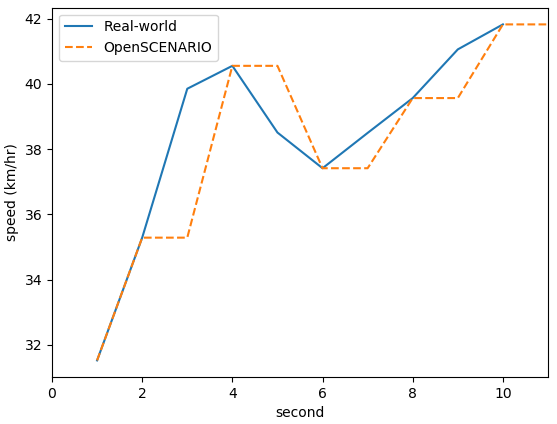}
         \caption{The figure shows the speed profile of the simulated trajectory that has been created using five samples of velocity and distance values.}
         \label{fig:speed_5_points}
    \end{subfigure}
    \caption{The figure depicts the different trajectory comparisons. It compares two different sample sizes of speed and distance parameters.}
    \label{fig:pos_comparison}
\end{figure}

We will use the scenario parameters to create concrete scenarios using the stochastic variation method in the extended work. It is necessary to consider meaningful and less number of parameters because more parameters can lead to the parameter-space explosion in scenario-based testing\cite{amersbach2019functional}. At the same time, the simulated scenario should reflect the real-world scenarios. In order to represent the real-world trajectory in simulation, we use speed and distance parameters. $m$ can have many parameters depending on the size we choose. So, it is important to select a meaningful number for $m$ to sample speed and distance values to represent the real-world trajectory. We have experimented with two different sample settings. The figure~\ref{fig:pos_comparison} shows the trajectories with two different $m$ values.  In the first sub figure~\ref{fig:pos_10_points}, ten samples of distance and speed values were used to recreate the trajectory of the adversary vehicle in a cut-in scenario. In the figure, it is clear that both real-world and simulated trajectories are similar. As the total duration of the scenario is ten seconds,  effectively, the sampling of speed and distance values is done at each second in this case. We have enabled the similarity check with the real-world and its simulated version using speed metric. Figure~\ref{fig:speed_10_points} depicts the comparison of speed profiles from real-world data and generated version.  Both figures~\ref{fig:pos_10_points} and~\ref{fig:speed_10_points} shows that the framework can extract scenarios from real-world data and recreate similar scenarios in simulation. We further down-sampled the number of speed and distance values to represent the trajectory to 5 as shown in the figure~\ref{fig:pos_5_points}. We also compare the speed profile as depicted in the figure~\ref{fig:speed_5_points}. Both figures show that the trajectory has slightly deviated from the real-world scenario and the speed profile is also not similar comparatively. We conclude from the whole comparison from the figure~\ref{fig:pos_comparison} that if we have a scenario spanning ten seconds, then consider at least five or more sampling values for $m$ to recreate the trajectory. Comparing different $m$ value settings is essential as it shows whether the selected threshold is enough for creating a similar trajectory. Thus it can influence the number of meaningful scenarios to reduce the effect of parameter space explosion in scenario-based testing.

\begin{table}[h!]
\vspace{2mm}
\centering
\caption{Scenario extraction accuracy of the framework}
\label{table:table_scenario_accuracy}
\begin{tabular}{|c c c c|} 
 \hline
 Scenario & Real-world & Generated & False +ve \\ [0.5ex] 
 \hline\hline
 Cut-in & 12 & 12 & 2\\ 
 \hline
 Cut-out & 6 & 6 & 0\\
 \hline
\end{tabular}
\end{table}

We have manually identified twelve cut-in and six cut-out scenarios in the data. The extraction framework generated all twelve cut-in scenarios with two false-positive cases. The framework extracts two false-positive cases while a vehicle was waiting at a junction, and gradually the vehicle joined the ego's lane from the road, that is, the left arm of the junction. In this case, the algorithm determined it as a cut-in event. The vehicle was not in the same lane as the ego vehicle and more than 1.5 meters away laterally, so the algorithm began tracking the vehicle. When they join the ego vehicle's lane in the intersection, it is marked as a cut-in scenario; however, these are not cut-in scenarios. We manually checked six cut-out scenarios in the data, and the framework could generate six of them. The result is shown in the table~\ref{table:table_scenario_accuracy}.

Once we generate scenarios in OpenX format, we can use them in any simulator that supports it. The external logic can control the motion of the Scenario object in OpenSCENARIO.  OpenSCENARIO supports the controller concept that enables external logic in generated OpenSCENRIO. If the external logic runs by SUT, we can evaluate the SUT performance in the generated scenario. We have added a controller for Carla\cite{dosovitskiy2017carla} in the generated OpenSCENARIO files. It enables running and assessing the SUT in CARLA.

\section{Discussion and Conclusion}
This paper proposes a novel scenario extraction framework for lane-change maneuvers. It uses lidar data to create a road network and then the object tracking information and road network to extract the lane change scenarios. The extracted scenarios and road networks are represented in OpenX formats. The proposed framework is an automated pipeline that can generate OpenX files without manually creating the content in the pipeline. The result shows that the framework can extract the real-world scenario and regenerate in a simulation. We are extending this work to build an urban validation dataset that has realistic behaviour of the traffic participants in the urban environment. Using the validation dataset, we can assess the SUT in realistic scenarios and generate many variations of concrete scenarios from the validation dataset to evaluate SUT in more dangerous scenarios.

The framework can be extended to other datasets with minimal modification in the dataset playback section. Many datasets provide point cloud with intensity, IMU, and tracking data, so we can extend the framework to other popular datasets by modifying the framework's playback section.

\bibliographystyle{IEEEtran}
\bibliography{main.bib}

\begin{thebibliography}{10}
\providecommand{\url}[1]{#1}
\csname url@samestyle\endcsname
\providecommand{\newblock}{\relax}
\providecommand{\bibinfo}[2]{#2}
\providecommand{\BIBentrySTDinterwordspacing}{\spaceskip=0pt\relax}
\providecommand{\BIBentryALTinterwordstretchfactor}{4}
\providecommand{\BIBentryALTinterwordspacing}{\spaceskip=\fontdimen2\font plus
\BIBentryALTinterwordstretchfactor\fontdimen3\font minus
  \fontdimen4\font\relax}
\providecommand{\BIBforeignlanguage}[2]{{%
\expandafter\ifx\csname l@#1\endcsname\relax
\typeout{** WARNING: IEEEtran.bst: No hyphenation pattern has been}%
\typeout{** loaded for the language `#1'. Using the pattern for}%
\typeout{** the default language instead.}%
\else
\language=\csname l@#1\endcsname
\fi
#2}}
\providecommand{\BIBdecl}{\relax}
\BIBdecl

\bibitem{standard2018j3016}
S.~Standard, ``J3016-taxonomy and definitions for terms related to driving
  automation systems for on-road motor vehicles,'' 2018.

\bibitem{koopman2019safety}
P.~Koopman, U.~Ferrell, F.~Fratrik, and M.~Wagner, ``A safety standard approach
  for fully autonomous vehicles,'' in \emph{International Conference on
  Computer Safety, Reliability, and Security}.\hskip 1em plus 0.5em minus
  0.4em\relax Springer, 2019, pp. 326--332.

\bibitem{koopman2016challenges}
P.~Koopman and M.~Wagner, ``Challenges in autonomous vehicle testing and
  validation,'' \emph{SAE International Journal of Transportation Safety},
  vol.~4, no.~1, pp. 15--24, 2016.

\bibitem{xinxin2020csg}
Z.~Xinxin, L.~Fei, and W.~Xiangbin, ``Csg: Critical scenario generation from
  real traffic accidents,'' in \emph{2020 IEEE Intelligent Vehicles Symposium
  (IV)}.\hskip 1em plus 0.5em minus 0.4em\relax IEEE, pp. 1330--1336.

\bibitem{zhao2016accelerated}
D.~Zhao, H.~Lam, H.~Peng, S.~Bao, D.~J. LeBlanc, K.~Nobukawa, and C.~S. Pan,
  ``Accelerated evaluation of automated vehicles safety in lane-change
  scenarios based on importance sampling techniques,'' \emph{IEEE transactions
  on intelligent transportation systems}, vol.~18, no.~3, pp. 595--607, 2016.

\bibitem{waymo2017road}
L.~Waymo, ``On the road to fully self-driving,'' \emph{Waymo Safety Report},
  pp. 1--43, 2017.

\bibitem{de2017assessment}
E.~de~Gelder and J.-P. Paardekooper, ``Assessment of automated driving systems
  using real-life scenarios,'' in \emph{2017 IEEE Intelligent Vehicles
  Symposium (IV)}.\hskip 1em plus 0.5em minus 0.4em\relax IEEE, 2017, pp.
  589--594.

\bibitem{enable2016enable}
E.-S. Consortium \emph{et~al.}, ``Enable-s3 european project,'' \emph{available
  at: www.enable-s3.eu}, 2016.

\bibitem{putz2017system}
A.~P{\"u}tz, A.~Zlocki, J.~Bock, and L.~Eckstein, ``System validation of highly
  automated vehicles with a database of relevant traffic scenarios,''
  \emph{situations}, vol.~1, pp. 19--22, 2017.

\bibitem{elrofai2016scenario}
H.~Elrofai, D.~Worm, and O.~O. den Camp, ``Scenario identification for
  validation of automated driving functions,'' in \emph{Advanced Microsystems
  for Automotive Applications 2016}.\hskip 1em plus 0.5em minus 0.4em\relax
  Springer, 2016, pp. 153--163.

\bibitem{amersbach2019functional}
C.~Amersbach and H.~Winner, ``Functional decomposition—a contribution to
  overcome the parameter space explosion during validation of highly automated
  driving,'' \emph{Traffic injury prevention}, vol.~20, no. sup1, pp. S52--S57,
  2019.

\bibitem{ponn2020identification}
T.~Ponn, M.~Breitfu{\ss}, X.~Yu, and F.~Diermeyer, ``Identification of
  challenging highway-scenarios for the safety validation of automated vehicles
  based on real driving data,'' in \emph{2020 Fifteenth International
  Conference on Ecological Vehicles and Renewable Energies (EVER)}.\hskip 1em
  plus 0.5em minus 0.4em\relax IEEE, 2020, pp. 1--10.

\bibitem{hauer2020clustering}
F.~Hauer, I.~Gerostathopoulos, T.~Schmidt, and A.~Pretschner, ``Clustering
  traffic scenarios using mental models as little as possible,'' in \emph{2020
  IEEE Intelligent Vehicles Symposium (IV)}.\hskip 1em plus 0.5em minus
  0.4em\relax IEEE, 2020, pp. 1007--1012.

\bibitem{najm2007pre}
W.~G. Najm, J.~D. Smith, M.~Yanagisawa \emph{et~al.}, ``Pre-crash scenario
  typology for crash avoidance research,'' United States. National Highway
  Traffic Safety Administration, Tech. Rep., 2007.

\bibitem{de2020real}
E.~De~Gelder, J.~Manders, C.~Grappiolo, J.-P. Paardekooper, O.~O. Den~Camp, and
  B.~De~Schutter, ``Real-world scenario mining for the assessment of automated
  vehicles,'' in \emph{2020 IEEE 23rd International Conference on Intelligent
  Transportation Systems (ITSC)}.\hskip 1em plus 0.5em minus 0.4em\relax IEEE,
  2020, pp. 1--8.

\bibitem{krajewski2018highd}
R.~Krajewski, J.~Bock, L.~Kloeker, and L.~Eckstein, ``The highd dataset: A
  drone dataset of naturalistic vehicle trajectories on german highways for
  validation of highly automated driving systems,'' in \emph{2018 21st
  International Conference on Intelligent Transportation Systems (ITSC)}.\hskip
  1em plus 0.5em minus 0.4em\relax IEEE, 2018, pp. 2118--2125.

\bibitem{tenbrock2021conscend}
A.~Tenbrock, A.~K{\"o}nig, T.~Keutgens, J.~Bock, H.~Weber, R.~Krajewski, and
  A.~Zlocki, ``The conscend dataset: Concrete scenarios from the highd dataset
  according to alks regulation unece r157 in openx,'' \emph{arXiv preprint
  arXiv:2103.09772}, 2021.

\bibitem{opendrive}
\BIBentryALTinterwordspacing
{Association for Standardization of Automation and Measuring Systems},
  ``{OpenDRIVE},'' 2021. [Online]. Available:
  \url{https://www.asam.net/standards/detail/opendrive/}
\BIBentrySTDinterwordspacing

\bibitem{openscenario}
\BIBentryALTinterwordspacing
------, ``{OpenSCENARIO},'' 2021. [Online]. Available:
  \url{https://www.asam.net/standards/detail/openscenario/}
\BIBentrySTDinterwordspacing

\bibitem{openstandar}
\BIBentryALTinterwordspacing
{MSC Software}, ``{Open Standards - Essential for Self-Driving?}'' 2021.
  [Online]. Available:
  \url{https://d10tcz9jtwksbg.cloudfront.net/wp-content/uploads/2020/02/Open-Standards-Essential-for-Self-Driving.pdf}
\BIBentrySTDinterwordspacing

\bibitem{werling2010optimal}
M.~Werling, J.~Ziegler, S.~Kammel, and S.~Thrun, ``Optimal trajectory
  generation for dynamic street scenarios in a frenet frame,'' in \emph{2010
  IEEE International Conference on Robotics and Automation}.\hskip 1em plus
  0.5em minus 0.4em\relax IEEE, 2010, pp. 987--993.

\bibitem{poggenhans2018lanelet2}
F.~Poggenhans, J.-H. Pauls, J.~Janosovits, S.~Orf, M.~Naumann, F.~Kuhnt, and
  M.~Mayr, ``Lanelet2: A high-definition map framework for the future of
  automated driving,'' in \emph{2018 21st International Conference on
  Intelligent Transportation Systems (ITSC)}.\hskip 1em plus 0.5em minus
  0.4em\relax IEEE, 2018, pp. 1672--1679.

\bibitem{dosovitskiy2017carla}
A.~Dosovitskiy, G.~Ros, F.~Codevilla, A.~Lopez, and V.~Koltun, ``Carla: An open
  urban driving simulator,'' in \emph{Conference on robot learning}.\hskip 1em
  plus 0.5em minus 0.4em\relax PMLR, 2017, pp. 1--16.

\end{thebibliography}

\end{document}